\documentclass[a4paper, 10pt, conference]{IEEEtran}
\IEEEoverridecommandlockouts
\usepackage{cite}
\usepackage{amsmath,amssymb,amsfonts}
\usepackage{algorithmic}
\usepackage{graphicx}
\usepackage{textcomp}
\usepackage{xcolor}
\usepackage{multirow}
\ifCLASSOPTIONcompsoc
    \usepackage[caption=false, font=normalsize, labelfont=sf, textfont=sf]{subfig}
\else
\usepackage[caption=false, font=footnotesize]{subfig}
\fi
\usepackage{balance}

\def\BibTeX{{\rm B\kern-.05em{\sc i\kern-.025em b}\kern-.08em
    T\kern-.1667em\lower.7ex\hbox{E}\kern-.125emX}}

\begin{document}

\title{A Memory Efficient Deep Reinforcement Learning Approach For Snake Game Autonomous Agents\\
\thanks{\textsuperscript{1}Research Assistant. 

\textsuperscript{2}Assistant Professor, IEEE Member.

\textsuperscript{*}GitHub implementation: https://github.com/rafattushar/rl-snake }}

\author{\IEEEauthorblockN{Md. Rafat Rahman Tushar\textsuperscript{1}}
\IEEEauthorblockA{\textit{Department of Electrical and Computer Engineering} \\
\textit{North South University}\\
Dhaka, Bangladesh \\
rafat.tushar@northsouth.edu}
\and
\IEEEauthorblockN{Shahnewaz Siddique\textsuperscript{2}}
\IEEEauthorblockA{\textit{Department of Electrical and Computer Engineering} \\
\textit{North South University}\\
Dhaka, Bangladesh \\
shahnewaz.siddique@northsouth.edu}
}

\maketitle


\begin{abstract}
To perform well, Deep Reinforcement Learning (DRL) methods require significant memory resources and computational time. Also, sometimes these systems need additional environment information to achieve a good reward. However, it is more important for many applications and devices to reduce memory usage and computational times than to achieve the maximum reward. This paper presents a modified DRL method that performs reasonably well with compressed imagery data without requiring additional environment information and also uses less memory and time. We have designed a lightweight Convolutional Neural Network (CNN) with a variant of the Q-network that efficiently takes preprocessed image data as input and uses less memory. Furthermore, we use a simple reward mechanism and small experience replay memory so as to provide only the minimum necessary information. Our modified DRL method enables our autonomous agent to play Snake, a classical control game. The results show our model can achieve similar performance as other DRL methods.\\

\end{abstract}



\section{Introduction}
Complex problems can be solved in real-world applications by carefully designing Deep Reinforcement Learning (DRL) models by taking high dimensional input data and producing discrete or continuous outputs. It is challenging to build a agent using sensory data capable of controlling and acting in an environment. The environment is also complex and primarily unknown to the acting agent. The agent needs to learn the underlying distribution of the state and action spaces, and the distribution changes as the agent encounters new data from an environment. Previously reinforcement learning algorithms \cite{q_learning92, td-gammon, pg_sutton_99, nat_ac, dpg} were presented with lower constraint problems to demonstrate the algorithms effectiveness. However, these systems were not well generalized for high dimensional inputs; thus, they could not meet the requirements of practical applications.

Recently, DRL has had success in CNN based vision-based problems \cite{atari, human-lavel, double_q}. They have successfully implemented DRL methods that learn to control based on image pixel. 
Although the image-based DRL methods have enjoyed considerable success, they are memory intensive during training as well as deployment. Since they require a massive amount of memory, they are not suitable for implementation in mobile devices or mid-range autonomous robots for training and deployment. 

All modern reinforcement learning algorithms use replay buffer for sampling uncorrelated data for online training in mainly off-policy algorithms. Experience replay buffer also improves the data efficiency \cite{replay_99} during data sampling. Since the use of neural networks in various DRL algorithms is increasing, it is necessary to stabilize the neural network with uncorrelated data. That is why the experience replay buffer is a desirable property of various reinforcement learning algorithms. The first successful implementation of DRL in high dimensional observation space, the Deep Q-learning \cite{atari}, used a replay buffer of $10^6$ size. After that, \cite{double_q, ddpg, maddpg, soft_ac}, to name a few, have solved complex high dimensional problems but still use a replay buffer of the same size.

Experience replay buffer suffers from two types of issues. One is to choose the size of the replay buffer, and the second is the method of sampling data from the buffer. 
\cite{priotized_replay, hindsight_replay, deeper_replay} consider the latter problem to best sample from the replay buffer. But the favorable size for the replay buffer remains unknown. Although \cite{deeper_replay} points out that the learning algorithm is sensitive to the size of the replay buffer, they have not come up with a better conclusion on the size of the buffer.    

In this paper, we tackle the memory usage of DRL algorithms by implementing a modified approach for image preprocessing and replay buffer size. Although we want the agent to obtain a decent score, we are more concerned about memory usage. 
We choose a Deep Q-Network (DQN) \cite{atari} for our algorithm with some variations. Our objective is to design a DRL model that can be implemented on mobile devices during training and deployment. To be deployed on mobile devices, memory consumption must be minimized as traditional DRL model with visual inputs sometimes need half a terabyte of memory. We achieve low memory consumption by preprocessing the visual image data and tuning the replay buffer size with other hyperparameters. Then, we evaluate our model in our simulation environment using the classical control game named Snake.\textsuperscript{*} 
The results show that our model can achieve similar performance as other DRL methods.

\section{Related Work}
The core idea of reinforcement learning is the sequential decision making process involving some agency that learns from the experience and acts on uncertain environments. After the development of a formal framework of reinforcement learning, many algorithms have been introduced such as, \cite{q_learning92, td-gammon, pg_sutton_99, nat_ac, dpg}. 

Q-learning \cite{q_learning92} is a model-free asynchronous dynamic programming algorithm of reinforcement learning. Q-learning proposes that by sampling all the actions in states and iterating the action-value functions repeatedly, convergence can be achieved. The Q-learning works perfectly on limited state and action space while collapsing with high dimensional infinite state space. Then, \cite{atari} proposes their Deep Q-network algorithm that demonstrates significant results with image data. Among other variations, they use a convolutional neural network and replay buffer. Double Q-learning \cite{double_q_10} is applied with DQN to overcome the overestimation of the action-value function and is named Deep Reinforcement Learning with Double Q-Learning (DDQN) \cite{double_q}. DDQN proposes another neural network with the same structure as DQN but gets updated less frequently. Refined DQN \cite{autonomous} proposes another DRL method that involves a carefully designed reward mechanism and a dual experience replay structure. Refined DQN evaluate their work by enabling their agent to play the snake game.  

The experience replay buffer is a desirable property of modern DRL algorithms. It provides powerful, model-free, off-policy DRL algorithms with correlated data and improves data efficiency \cite{replay_99} during data sampling. DQN \cite{atari} shows the power of replay buffer in sampling data. DQN uses the size $10^6$ for replay buffer. After that, \cite{double_q, autonomous, ddpg, maddpg, soft_ac}, among others, have shown their work with the same size and structure as the replay buffer. Schaul et al. propose an efficient sampling strategy in their prioritized experience replay (PER) \cite{priotized_replay}. PER shows that instead of sampling data uniform-randomly, the latest data gets the most priority; hence the latest data have more probability of being selected, and this selection method seems to improve results. \cite{deeper_replay} shows that a large experience replay buffer can hurt the performance. They also propose that when sampling data to train DRL algorithms, the most recent data should the appended to the batch.

\section{Method}

Our objective is to reduce memory usage during training time while achieving the best performance possible. The replay memory takes a considerable amount of memory, as described later. We try to achieve memory efficiency by reducing the massive replay buffer requirement with image preprocessing and the buffer size. The buffer size is carefully chosen so that the agent has the necessary information to train well and achieves a moderate score. We use a slight variation of the deep Q-learning algorithm for this purpose.

\begin{table}[!t]
\renewcommand{\arraystretch}{1.3}
\caption{Reward Mechanism for Snake Game}
\centering
\begin{tabular}{ccc}
\hline
\textbf{Moves}&\textbf{Rewards}&\textbf{Results} \\
Eats an apple & +1 & Score Increase \\
Hits with wall or itself & -1 & End of episode \\
Not eats or hits wall or itself & -0.1 & Continue playing games \\
\hline
\end{tabular}
\label{tab_reward_snake}
\end{table}

\begin{table}[!t]
\renewcommand{\arraystretch}{1.3}
\caption{Memory Requirement for Different Pixel Data}
\centering
\begin{tabular}{cccc}
\hline
\textbf{}&\textbf{RGB}&\textbf{Grayscale}&\textbf{Binary} \\
\textbf{Data Type} & float & float & int \\
\textbf{Size (kB)} & 165.375 & 55.125 & 6.890 \\
\textbf{Memory Save \% w.r.t. RGB} & 0\% & 67\% & 96\% \\
\textbf{Memory Save \% w.r.t. Grayscale} & - & 0\% & 87.5\% \\
                     

\hline
\end{tabular}
\label{tab_image_memory}
\end{table}


\subsection{Image Preprocessing}\label{preproscessing}

\begin{figure}[!t]
\centering
\subfloat[Before preprocessing]
{\includegraphics[trim={2.5cm 0 2.5cm 0},clip,width=1.5in]{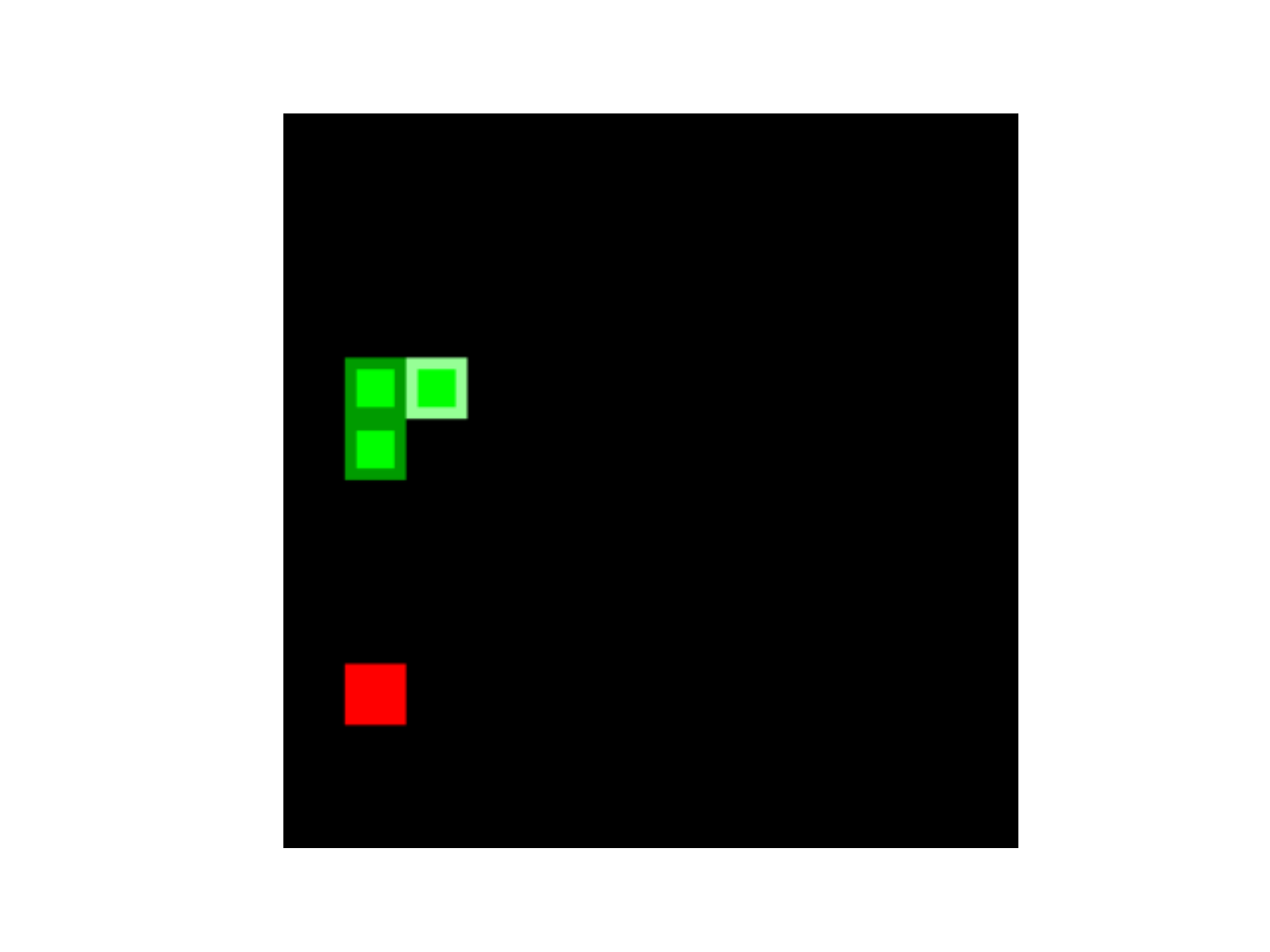}
\label{prepros_before}}
\hfil
\subfloat[After preprocessing]
{\includegraphics[trim={2.5cm 0 2.5cm 0},clip,width=1.5in]{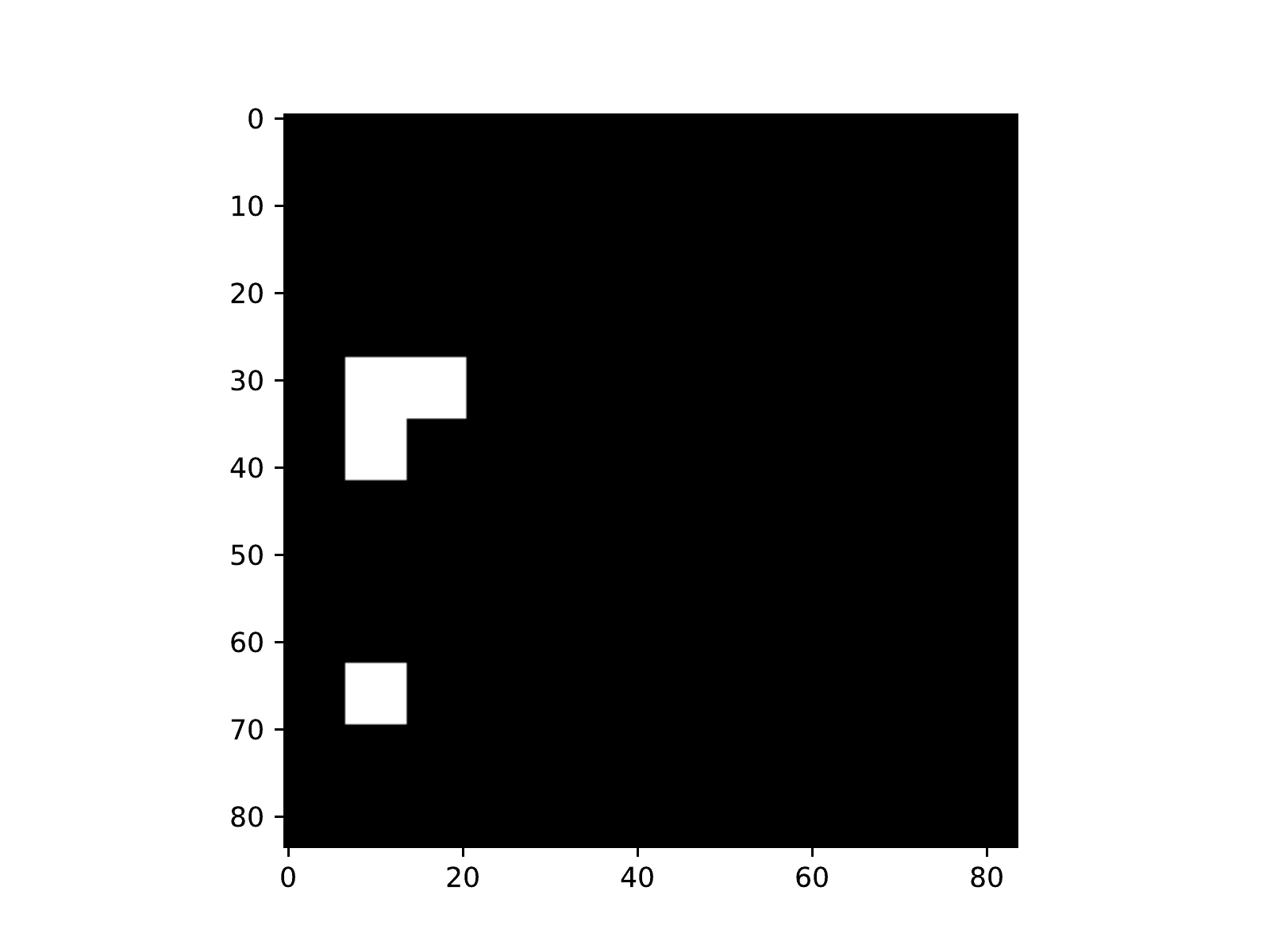}
\label{prepros_after}}
\caption{Visual image data before and after preprocessing}
\label{prepros}
\end{figure}

\begin{figure}[!b]
    \centering
    \includegraphics[width=\linewidth]{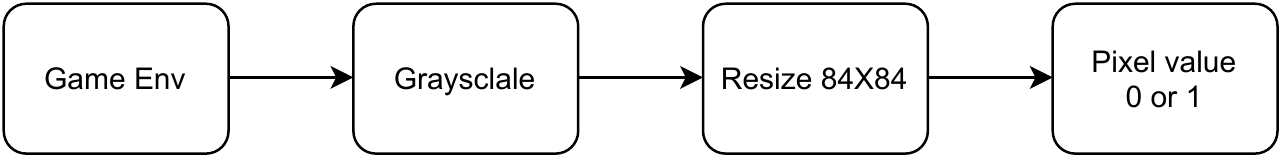}
    \caption{Diagram of image preprocessing}
    \label{diagram_prepros}
\end{figure}

The agent gets the RGB values in the 3-D array format from the games' environments. We convert the RGB array into grayscale because it would not affect the performance~\cite{image-colorization} and it saves three times of memory. We resize the grayscale data into $84\times84$ pixels. Finally, for more memory reduction, we convert this resized grayscale data into binary data (values only with 0 and 1). The memory requirement for storing various image data (scaled-down between 0 and 1) is given in Table~\ref{tab_image_memory}. Table~\ref{tab_image_memory} shows that it saves around 67\% from converting RGB into grayscale and around 96\% from converting RBG into binary. Also, the memory requirement reduces by around 87.5\% converting from grayscale into binary. Visual pixel data transformation with preprocessing is given in Fig. \ref{prepros}. The preprocessing method is presented using a flowchart in Fig.~\ref{diagram_prepros}. 

\subsection{Game Selection and Their Environments}

The use-case of our target applications is less complex tasks. For this reason, we implemented the classical Snake game \cite{snake} in the 'pygame' module. The game screen is divided into a $12\times12$ grid. The resolution for the game is set to $252\times252$. The initial snake size is 3. The controller has four inputs to navigate. Table~\ref{tab_reward_snake} shows the valid actions and respective reward for the snake game environment.

\subsection{Reinforcement Learning Preliminary}

Any reinforcement learning or sequential decision-making problem can be formulated with Markov Decision Processes (MDPs). An MDP is a triplet $M = ( \mathcal{X, A, P}_0)$, where $\mathcal{X}$ is a set of valid states, $\mathcal{A}$ is a set of valid actions, and $\mathcal{P}_0$ is transition probability kernel that maps $\mathcal{X \times A}$ into next state transition probability. For a deterministic system, the state transition is defined as,
\begin{equation}
    \label{eq-state_transition}
    s_{t+1} = f(s_t, a_t)
\end{equation}
The reward is defined as,
\begin{equation}
    \label{eq-reward_transition}
    r_t = R(s_t, a_t)
\end{equation}
The cumulative reward over a trajectory or episode is called the return, $R(\tau)$. The equation for discounted return is given below,
\begin{equation}
    \label{eq-discounted_return}
    R(\tau) = \sum_{t=0}^{\infty} \gamma^t r_t
\end{equation}

\subsection{Deep Q-Learning}

The goal of the RL agent is to maximize the expected return.  Following a policy $\pi$, the expected return, $J(\pi)$, is defined as,
\begin{equation}
    \label{eq-expected_return}
    J(\pi) = \underset{\tau \sim \pi}{\mathbb{E}} [R(\tau)]
\end{equation}
The optimal action-value or q function $Q^*(s, a)$ maximizes the expected return by taking any action at state $s$ and acting optimally in the following states.
\begin{equation}
    \label{eq-optimal_action_value}
    Q^*(s,a) = \max_{\pi} \underset{\tau \sim \pi}{\mathbb{E}} [R(\tau) | s_0 = s, a_0 = a]
\end{equation}
For finding out the optimal actions based on an optimal action-value function at time $t$, the $Q^*$ must satisfy the Bellman Equation, which is,
\begin{equation} 
    \label{eq-bellman_op_action_value}
    Q^*(s,a) = \underset{s' \sim \rho}{\mathbb{E}} \left[ r(s,a) + \gamma \max_{a'} Q^*(s', a') \right]
\end{equation}
The optimal action-value function gives rise to optimal action $a^*(s)$. The $a^*(s)$ can be described as,
\begin{equation}
    \label{eq-optimal_action}
    a^*(s) = \arg \max_{a} Q^*(s,a)
\end{equation}
For training an optimal action-value function, sometimes a non-linear function approximator like neural network~\cite{atari} is used. We used a convolutional neural network.

\begin{table}[!t]
\caption{The architecture of Neural Network}
\centering
\begin{tabular}{ccccccc}
\hline
\textbf{Layer}&\textbf{Filter}&\textbf{Stride}&\textbf{Layer}&\textbf{Acti-}&\textbf{Zero}&\textbf{Output} \\
\textbf{Name} & & & &\textbf{vation} &\textbf{Padd} & \\
Input & & & & & & 84*84*4 \\
Conv1 &	8*8	& 4 & 32 & ReLU & Yes & 21*21*32 \\
M. Pool & 2*2 & 2 & & & Yes & 11*11*32 \\
Conv2 & 4*4 & 2 & 64 & ReLU & Yes & 6*6*64 \\
M. Pool & 2*2 & 2 & & & Yes & 3*3*64 \\
B. Norm &  &  &  &  &  & 3*3*64 \\
Conv3 & 3*3 & 2 & 128 & ReLU & Yes & 2*2*128 \\
M. Pool & 2*2 & 2 & & & Yes & 1*1*128 \\
B. Norm &  &  &  &  &  & 1*1*128 \\
Flatten & & & & & & 128 \\
FC & & & 512 & ReLU & & 512 \\
FC & & & 512 & ReLU & & 512 \\
Output & & &No. of & Linear & &No. of \\
& & &actions & & & actions\\
\hline
\multicolumn{7}{l}{\textsuperscript{M. Pool = Max Pooling, B. Norm = Batch Normalization, FC = Fully Connected}}
\end{tabular}
\label{tab_neural}
\end{table}

\begin{table}[!t]
\caption{Memory Requirement Experience Replay}
\centering
\begin{tabular}{cccc}
\hline
\textbf{}&\textbf{RGB}&\textbf{Grayscale}&\textbf{Binary} \\
\textbf{Memory Usage (GB)} & 1261.71 & 420.57 & 2.628 \\
\textbf{Memory Save \% w.r.t. RGB} & 0\% & 67\% & 99.7\% \\
\textbf{Memory Save \% w.r.t. Grayscale} & - & 0\% & 99.4\% \\
                     

\hline
\end{tabular}
\label{tab_replay_memory}
\end{table}

\begin{figure}[!b]
    \centering
    \includegraphics[trim={0 15cm 3cm 4cm},clip,width=\linewidth, height=5cm]{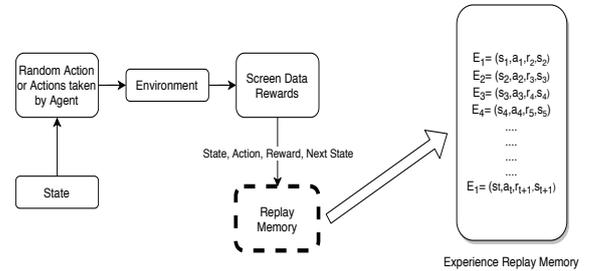}
    \caption{Structure of experience replay memory and flowchart}
    \label{replay_mem}
\end{figure}

\begin{figure*}[t]
    \centering
    \includegraphics[trim={0 12cm 5cm 3cm},clip,height=0.70\textwidth, width=\textwidth]{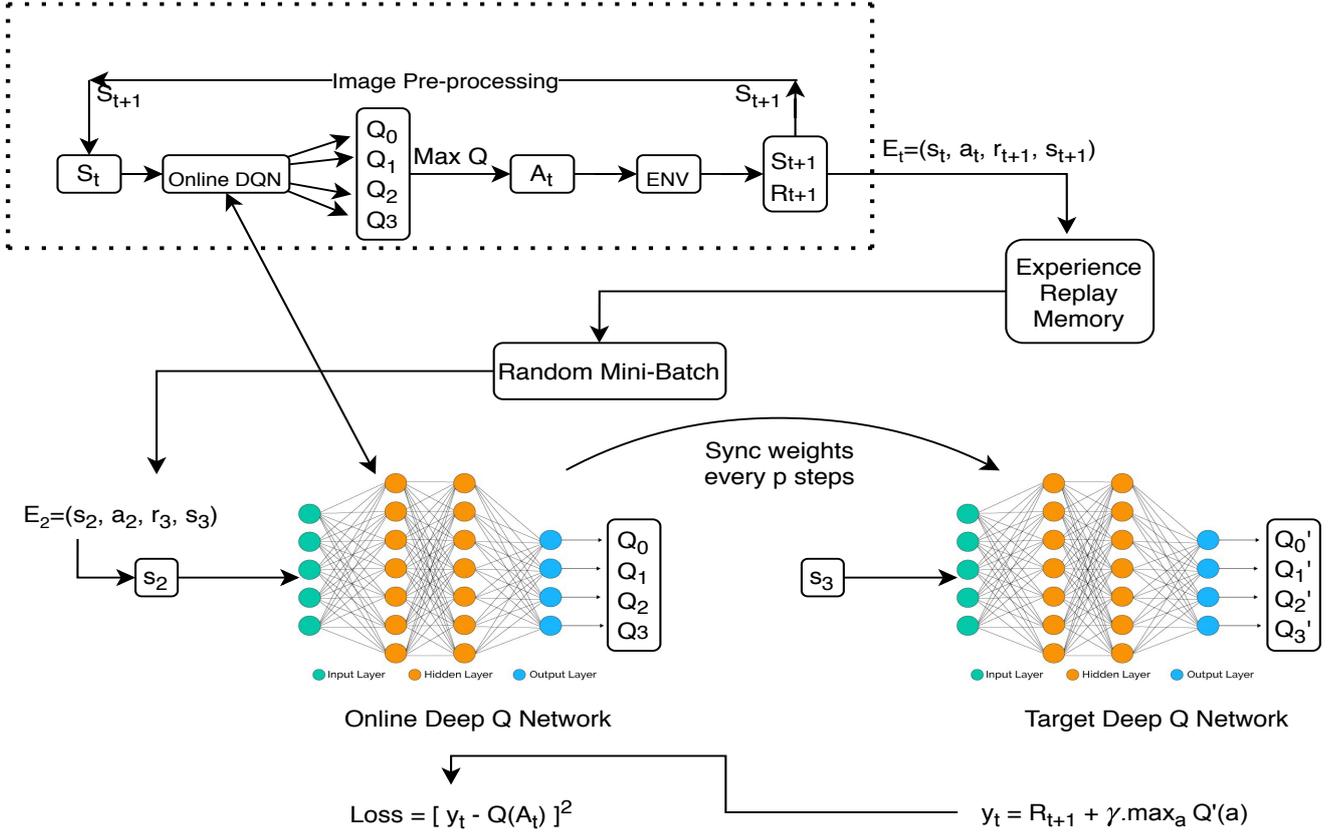}
    \caption{The deep reinforcement learning design structure of our model}
    \label{full_design}
\end{figure*}

\subsection{Neural Network}

The action-value function is iteratively updated to achieve the optimal action-value function. The neural network used to approximate the action-value function and update at each iteration is called Q-network. We train the Q-network, parameterized by $\theta$, by minimizing a loss function $L_i(\theta_i)$ at $i th$ iteration.
\begin{equation}
    \label{eq-loss_function}
    L_i(\theta_i) = \underset{s,a \sim \rho}{\mathbb{E}} \left[ (y_i - Q(s,a; \theta_i))^2 \right]
\end{equation}
where $y_i = \underset{s' \sim \rho}{\mathbb{E}} \left[ r(s,a) + \gamma \underset{a'}{\max}  Q'(s', a';\theta'_{k}) \right]$ is the target for that update. Here $Q'$ is another Q-network with the same shape as Q-network but with a frozen parameter called target Q-network for training stability parameterized by $\theta'_k$. We train the Q-network by minimizing this loss function~(\ref{eq-loss_function}) w.r.t. the parameter $\theta_i$. We use Adam~\cite{adam} optimizer for fast convergence. Our convolutional neural network structure is shown in Table \ref{tab_neural}.

\subsection{Experience Replay Buffer}\label{replay}

As our focus is to keep memory requirements as low as possible during training, choosing the size of the replay buffer is one of the critical design decisions. The size of the replay buffer directly alters the requirement of memory necessity. We use a replay buffer of size 50,000, requiring less memory (only 5\%) than \cite{atari, double_q, autonomous}, which use a replay buffer of size 1,000,000. \cite{atari, double_q, autonomous} store grayscale data into a replay buffer. Table \ref{tab_replay_memory} shows that we use 99.4\% less memory compared to these works. The replay buffer stores data in FIFO (first in, first out) order so that the buffer contains only the latest data. We present the complete cycle of the experience replay buffer in Fig \ref{replay_mem}. Fig. \ref{full_design} illustrates our complete design diagram.  



\section{Experiments}


\subsection{Training}
For training our model, we take a random batch of 32 experiences from the replay buffer at each iteration. Our model has two convolutional neural networks (online DQN and target DQN) sharing the same structure but does not sync automatically. The weights of the target network are frozen so that it cannot be trained. The state history from the mini-batch is fed into the Online DQN. The DQN outputs the Q-values, $Q(s_t,a_t)$.
\begin{equation}
    \label{eq6}
    Loss=[y_t-Q(s_t,a_t)]^2
\end{equation}
The $y_t$ is calculated from the target Q-network. We are passing the next-state value to the target Q-network, and for each next-state in the batch, we get Q-value, respectively. That is our $max_{a'}Q(s',a')$ value in the below equation.
\begin{equation}
    \label{eq7}
    y_t=R_{t+1}+\gamma max_{a'} Q(s',a')
\end{equation}
The $\gamma$ is the discount factor, which is one of many hyperparameters we are using in our model. Initially, we set $\gamma$ value to 0.99. The $R_{t+1}$ is the reward in each experience tuple. So, we get the $y_t$ value. The loss function is generated by putting these values in \eqref{eq6}. Then, we use this loss function to backpropagate our Online DQN with an ‘Adam’ optimizer. Adam optimizer is used instead of classical stochastic gradient descent for more speed. The target DQN is synced with online DQN at every 10,000 steps. The values of hyperparameters we choose are listed in Table~\ref{tab_hyperparameter}.

\subsection{Results and Comparisons}
\begin{figure}[!t]
\centering
\subfloat[Score vs. episode graph]
{\includegraphics[width=0.48\linewidth]{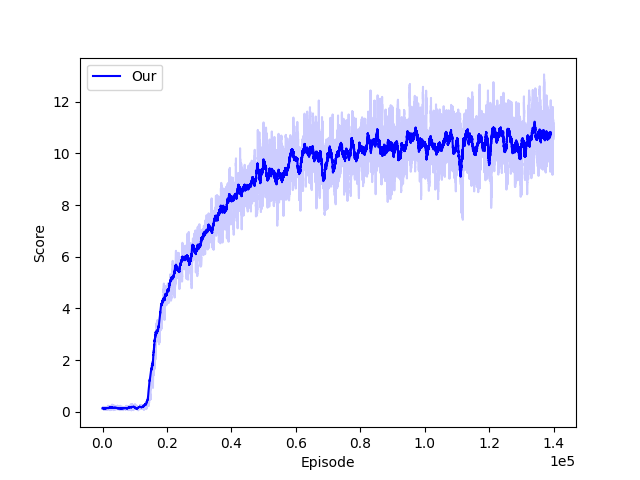}
\label{fig-res_our_score}}
\subfloat[Reward vs. episode graph]
{\includegraphics[width=0.48\linewidth]{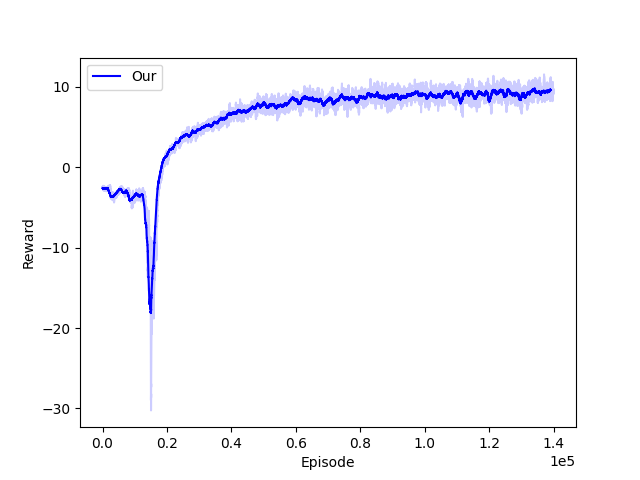}
\label{fig-res_our_reward}}
\caption{Results of our agent playing Snake game during training}
\label{fig-res_our}
\end{figure}

\begin{figure}[!t]
\centering
\subfloat[Score vs. episode graph]
{\includegraphics[width=0.48\linewidth]{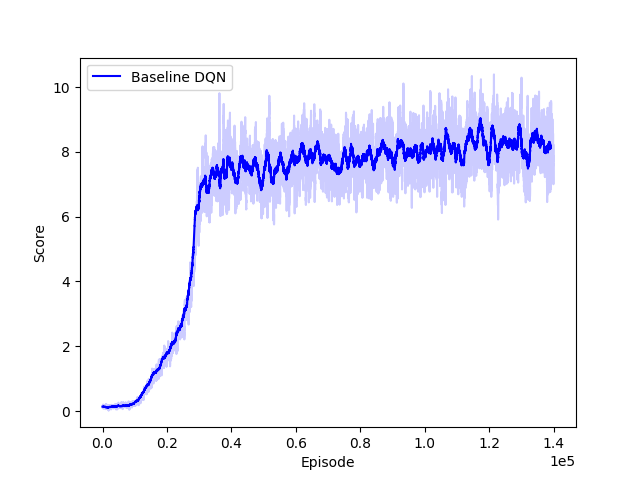}
\label{fig-res_baseline_score}}
\subfloat[Reward vs. episode graph]
{\includegraphics[width=0.48\linewidth]{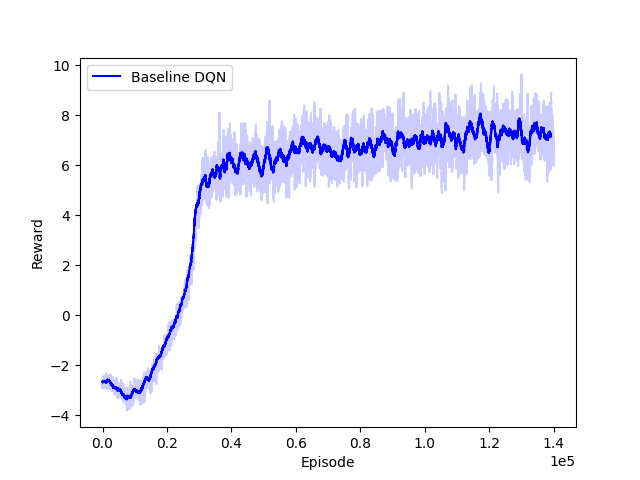}
\label{fig-res_baseline_reward}}
\caption{Results of baseline DQN model playing Snake game during training}
\label{fig-res_baseline}
\end{figure}

\begin{figure}[!b]
\centering
\subfloat[Score comparison]
{\includegraphics[width=0.48\linewidth]{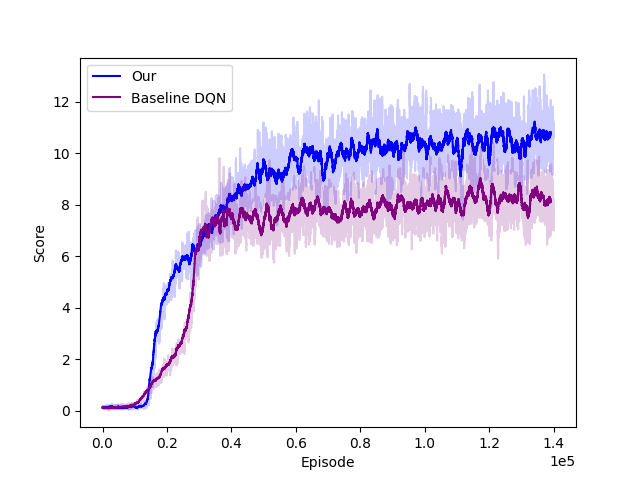}
\label{fig-res_compare_score}}
\subfloat[Reward comparison]
{\includegraphics[width=0.48\linewidth]{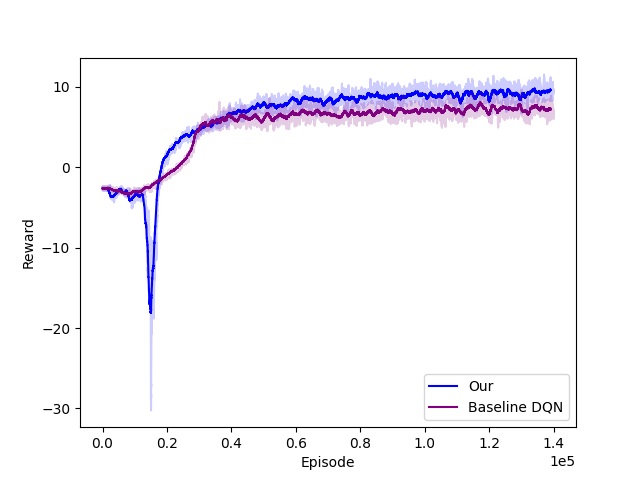}
\label{fig-res_compare_reward}}
\caption{Comparison between our model and baseline DQN model}
\label{fig-res_compare}
\end{figure}

We allow DRL agents to play 140,000 episodes of games to match the training results presented in \cite{autonomous}. We train one agent with our method and another with the DQN method presented in \cite{atari}, we refer to \cite{atari} as the baseline DQN model. Next, we compare our model with the baseline DQN model \cite{atari} and the refined DQN model \cite{autonomous}. The results of training the snake game with our model are shown in Fig. \ref{fig-res_our}. Fig. \ref{fig-res_our}\subref{fig-res_our_score} shows the game's score with our model during training. Fig. \ref{fig-res_our}\subref{fig-res_our_reward} shows that even though our reward mechanism is simpler than the refined DQN model, the agent maximizes the cumulative reward optimally.  

In section \ref{replay} we showed that our model is more memory efficient than the baseline DQN model and the refined DQN model during training. In this section we show that despite low memory usage, our model can achieve similar if not better results than the baseline and refined DQN models. Fig. \ref{fig-res_baseline} displays the baseline DQN results during training on the snake game. In Fig. \ref{fig-res_compare} we present the score and reward comparison between our model and the baseline DQN model. The blue line in Fig. \ref{fig-res_compare}\subref{fig-res_compare_score} represents our model's score, and the purple line represents the score of the baseline DQN model. During 140,000 numbers of training episodes, our model remains better at episode score though it requires fewer resources. Fig. \ref{fig-res_compare}\subref{fig-res_compare_reward} demonstrates that our model is capable of achieving higher cumulative rewards than the baseline DQN model. 

\begin{figure}[!t]
\centering
\subfloat[Score graph of Refined DQN (graph taken from~\cite{autonomous})]
{\includegraphics[trim={0 0 0 0},clip,width=0.47\linewidth]{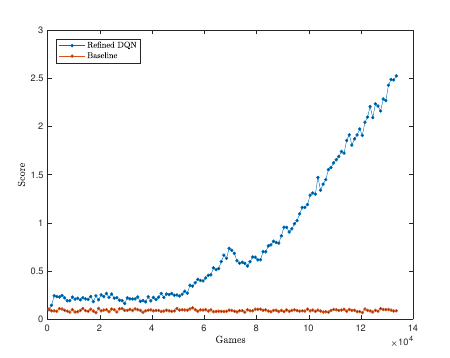}
\label{our_and_refined_refined}}
\subfloat[Score graph of our model]
{\includegraphics[height=3.5cm,width=0.47\linewidth]{ScoreEpisodeOur.png}
\label{our_and_refined_our}}
\caption{Comparison between Refined DQN model and our model}
\label{our_and_refined}
\end{figure}

\begin{figure}[!t]
\centering
\subfloat[Refined DQN score (Taken from~\cite{autonomous})]
{\includegraphics[trim={0 0 0 0},clip,width=0.48\linewidth]{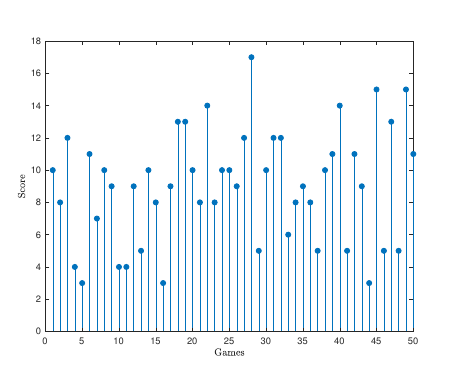}
\label{our_and_refined_test_refined}}
\hfil
\subfloat[Our model’s score]
{\includegraphics[trim={0 0 0 1cm},clip,height=3.25cm,width=0.48\linewidth]{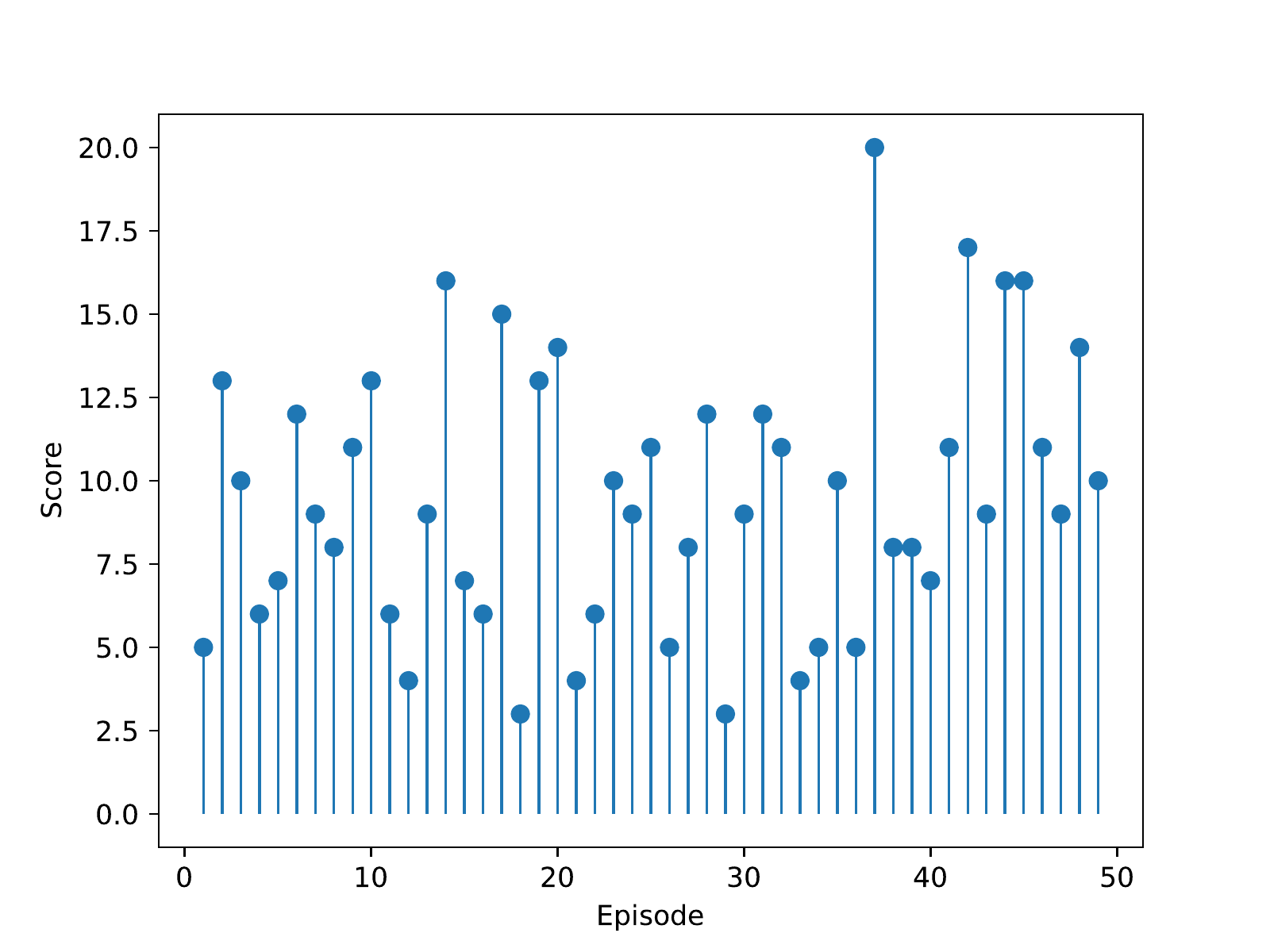}
\label{our_and_refined_test_our}}
\caption{Testing evaluation by playing random 50 episodes game}
\label{our_and_refined_test}
\end{figure}

\begin{table}[!b]
\caption{List of Performance comparison of Different Agents}
\centering
\begin{tabular}{cc}
\hline
\textbf{Performance}&\textbf{Score} \\
Human Average    & 1.98 *\\
Baseline Average    & 0.26 *\\
Refined DQN Average    & 9.04 *\\
\textbf{Our Average} &\textbf{9.53} \\
Human Best    & 15 *\\
Baseline Best    & 2 *\\
Refined DQN Best    & 17 *\\
\textbf{Our Best} &\textbf{20}\\
\hline
\multicolumn{2}{l}{* Data taken from~\cite{autonomous}}
\end{tabular}
\label{tab_performance_list}
\end{table}

We also compare the results between our model and the refined DQN model \cite{autonomous}. Refined DQN follows a dual experience replay memory architecture and a complex reward mechanism. However, our model surpasses their score. Since their game is similar to ours, we compare our results with the results provided in their paper. Fig.~\ref{our_and_refined}\subref{our_and_refined_refined} shows the results presented in \cite{autonomous}, and Fig.~\ref{our_and_refined}\subref{our_and_refined_our} is our model's results during training. By comparing Fig.~\ref{our_and_refined}\subref{our_and_refined_refined} and Fig.~\ref{our_and_refined}\subref{our_and_refined_our}, we can safely say that our model achieves better scores despite having a simple replay buffer, a simple reward mechanism, and less memory consumption. 

Fig.~\ref{our_and_refined_test}\subref{our_and_refined_test_refined} and Fig.~\ref{our_and_refined_test}\subref{our_and_refined_test_our} show scores of random 50 episodes during testing of refined DQN and our model, respectively. Table \ref{tab_performance_list} summarizes the scores provided in the refined DQN and our model. We can identify from Table \ref{tab_performance_list} that their refined DQN average is 9.04, while ours is 9.53, and their refined DQN best score is 17, while ours is 20. So, we can see that our model also performs better in the training and testing phase.

\begin{table}[ht]
\renewcommand{\arraystretch}{1.3}
\caption{List of Hyperparameters}
\centering
\begin{tabular}{ccc}
\hline
\textbf{Hyperparameter}&\textbf{Value}&\textbf{Description} \\
Discount Factor & 0.99 & $\gamma$-value in max Q-function \\
Initial Epsilon	& 1.0 & Exploration epsilon initial value \\
Final Epsilon & 0.01 & Exploration final epsilon value \\
Batch size & 32 & Mini batch from replay memory \\
Max step & 10,000 & Maximum number of steps \\
        &       & allowed per episode \\
Learning Rate & 0.0025 & Learning rate for Adam optimizer \\
Clip-Norm & 1.0 & Clipping value for Adam optimizer \\
Random Frames & 50,000 & Number of random initial steps \\
Epsilon greedy & 500,000 & Number of frames in which initial \\
frames      &           & epsilon will be equal final epsilon \\
Experience Replay & 50,000 & Capacity of experience replay \\
Memory          &       & memory \\

Update of DQN & 4 & The number of steps after each \\
            &   & update of DQN takes place \\
Update Target & 10,000 & The number of steps after the \\
DQN         &       & Target and Online DQN sync \\
\hline
\end{tabular}
\label{tab_hyperparameter}
\end{table}

\section{Conclusion}

In this paper, we have shown that better image preprocessing and constructing a better mechanism for replay buffer can reduce memory consumption on DRL algorithms during training. We have also demonstrated that using our method, the performance of the DRL agent on a lower constraint application is entirely similar, if not better. We combined our method with the DQN (with some modification) algorithm to observe the method's effectiveness. Our presented design requires less memory and a simple CNN. We established that our method's result is as good as other DRL approaches for the snake game autonomous agent. 


\section*{Acknowledgment}

This work was supported by North South University research grant CTRG-21-SEPS-18.


The authors would like to gratefully acknowledge that the computing resources used in this work was housed at the National University of Sciences and Technology (NUST), Pakistan. The cooperation was pursued under the South Asia Regional Development Center (RDC) framework of the Belt \& Road Aerospace Innovation Alliance (BRAIA).

\bibliographystyle{IEEEtran}

\bibliography{bibl}

\end{document}